\documentclass[final]{cvpr}

\usepackage{times}
\usepackage{epsfig}
\usepackage{graphicx}
\usepackage{amsmath}
\usepackage{amssymb}
\usepackage{epsfig}
\usepackage{graphicx}
\usepackage{amsmath}
\usepackage{amssymb}
\usepackage{times}
\usepackage{epsfig}
\usepackage{graphicx}
\usepackage{times}
\usepackage{epsfig}
\usepackage{graphicx}
\usepackage{amsmath}
\usepackage{amssymb}
\usepackage{multirow}
\usepackage[normalem]{ulem}
\usepackage{comment}
\usepackage{sidecap}
\usepackage[dvipsnames]{xcolor}
\usepackage{booktabs}
\usepackage{tabularx} 
\usepackage{algorithm,algorithmicx,algpseudocode}
\usepackage{caption,subcaption}
\usepackage{multicol}

\usepackage[pagebackref=true,breaklinks=true,colorlinks,bookmarks=false]{hyperref}

\usepackage{amsthm}
\newtheorem*{remark}{Remark}

\def\cvprPaperID{1954} 
\def\confYear{CVPR 2021}

\makeatletter
\newcommand{\printfnsymbol}[1]{%
  \textsuperscript{\@fnsymbol{#1}}%
}

\begin{document}

\pagenumbering{gobble}
\title{Semantic-aware Knowledge Distillation for Few-Shot\\ Class-Incremental Learning}

\author{ Ali Cheraghian\thanks{denotes equal contribution.} $^{,1,2}$, Shafin Rahman\printfnsymbol{1}$^{,3}$, Pengfei Fang$^{1,2}$, Soumava Kumar Roy$^{1,2}$\\ Lars Petersson$^{1,2}$, Mehrtash Harandi$^{2,4}$ \\

$^{1}$Australian National University, $^{2}$Data61-CSIRO, Australia\\ $^{3}$ North South University, Bangladesh, $^{4}$Monash University, Australia\\
{\tt\small \{Ali.Cheraghian,Pengfei.Fang,~Soumava.KumarRoy\}@anu.edu.au,}\\
{\tt\small shafin.rahman@northsouth.edu, Lars.Petersson@data61.csiro.au, mehrtash.harandi@monash.edu}
}


\maketitle

\begin{abstract}


Few-shot class incremental learning (FSCIL) portrays the problem of learning new concepts gradually, where only a few examples per concept are available to the learner. Due to the limited number of examples for training, the techniques developed for standard incremental learning cannot be applied verbatim to FSCIL.  In this work, we introduce a distillation algorithm to address the problem of FSCIL and propose to make use of  semantic information during training. To this end, we make use of word embeddings as semantic information which is cheap to obtain and which facilitate the distillation process. Furthermore, we propose a method based on an attention mechanism on multiple parallel embeddings of visual data to align visual and semantic vectors, which reduces issues related to catastrophic forgetting. Via experiments on MiniImageNet, CUB200, and CIFAR100 dataset, we establish new state-of-the-art results by outperforming existing approaches.


\end{abstract}

\section{Introduction}

\begin{figure}[!t]
\includegraphics[width=1\linewidth,trim=0cm 0cm 0cm 0cm, clip]{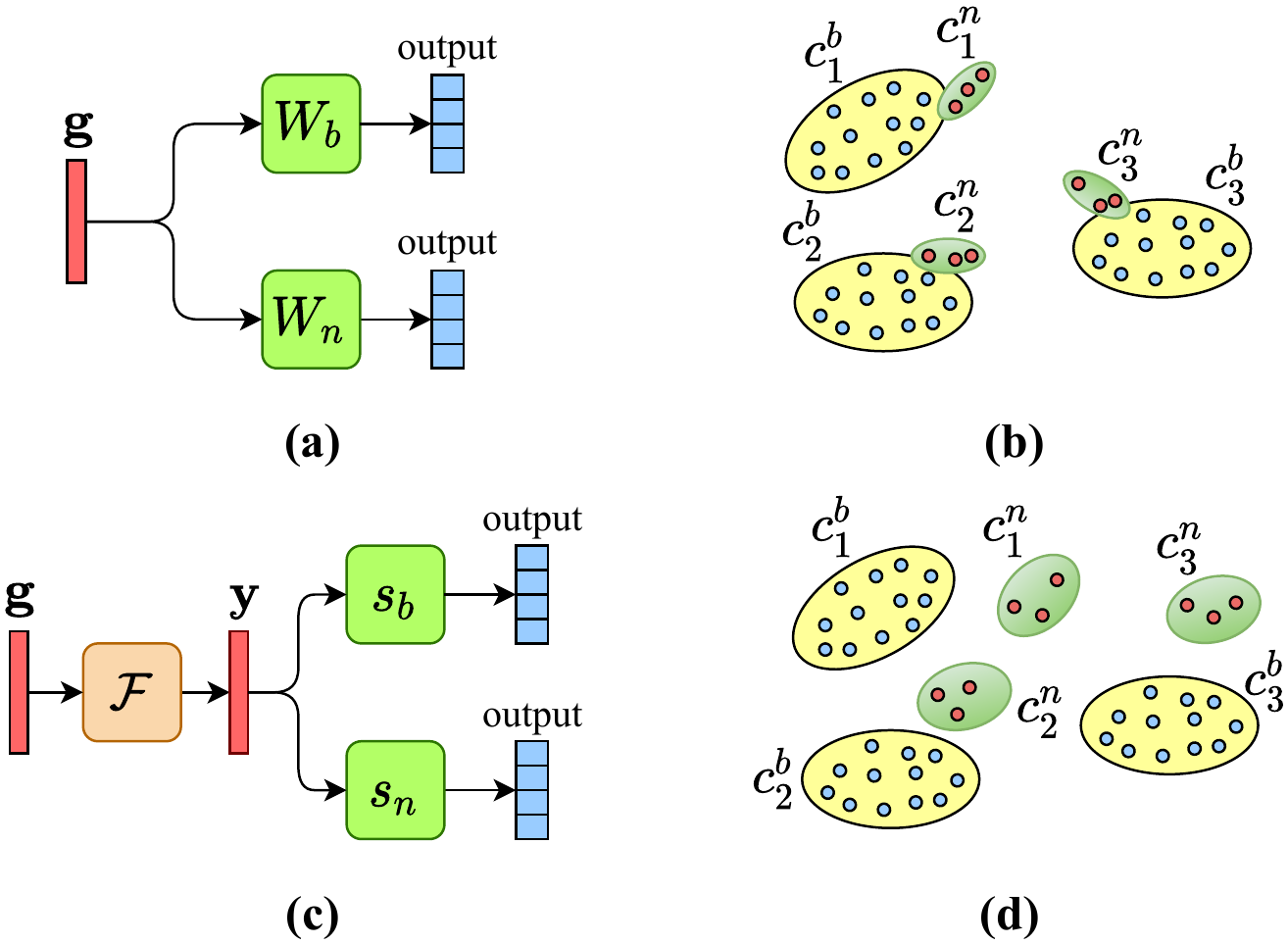}
\centering
\caption{
\small (a) Knowledge distillation as described in~\cite{8107520} does not work on few-shot class-incremental learning~\cite{Tao_2020_CVPR} since adding new tasks appends new trainable weights ($W_n$) to the network in addition to base weights ($W_b$). (b) The impact of using only a few instances of novel classes. As few samples are not sufficient to learn new parameters, the network gets biased towards base classes, overfitted on few examples of novel classes, and not well-separated from base classes. (c) Our semantically guided network does not add new parameters while adding new classes incrementally. We only include word vectors of new tasks ($s_n$) in addition to the base classes ($s_b$) and keep fine-tuning the base network ($\mathcal{F}$) (d) As a result, the knowledge distillation process can help the network, remembering base training, generalizing to novel classes, and finding well-separated representation of classes.
}
\label{fig:architecture_diff}
\vspace{-0.8cm}
\end{figure}

In a real world scenario, we may not get access to information about all possible classes when the system is first trained. It is more realistic to assume that we will obtain class-specific data incrementally as time goes by. Therefore, in such a scenario, we require that our model can be adapted with new information made available without hampering the performance on what has been learnt so far.  Although a natural task for human beings, it is a difficult task for an intelligent machine due to the possibility of catastrophic forgetting~\cite{MCCLOSKEY1989109}. A trained model tends to forget old tasks when learning new information. There are three different streams of work in the literature addressing such an incremental or continual learning paradigm ~\cite{DBLP:journals/corr/abs-1904-07734}. Firstly, {\it task-incremental learning} divides all classes into different tasks, where each task contains a few classes, and then learns each task individually. The task labels of the test instances are made available during testing which means the model does not need to predict the correct label between all classes but only between classes that are defined for a specific task. Secondly, {\it domain-incremental learning} does not reveal the task label at test time, but the model always solves the current task at hand without inferring the true class label.
Thirdly, {\it class-incremental learning} predicts the class label between all classes during test time as the output of all tasks are merged into one unified classifier without having access to the task label. Being the most realistic of the three, in this paper, we are interested in this third setting. Furthermore, in many applications, new tasks (a set of novel classes) come with only a few examples per class, making the class-incremental learning even more challenging. This setting is called {\it few-shot class-incremental learning} (FSCIL)~\cite{Tao_2020_CVPR}. 
The main challenges in FSCIL are catastrophic forgetting of already acquired knowledge and overfitting the network to novel classes. Challenges of that nature are addressed by the work on {\it knowledge distillation} in~\cite{44873}. However, \cite{Tao_2020_CVPR} showed that knowledge distillation is not the preferred approach for FSCIL due to class imbalance in the few-shot scenario and the performance trade-off between novel classes and base classes. In this paper, we propose an augmented knowledge distillation approach suitable for the case of few-shot incremental learning.


In order to apply knowledge distillation to novel tasks, scores of the previously trained model are needed as well as many instances of the new classes to be learned. Those new instances help to learn the new trainable weights that are added while learning novel tasks. For incremental learning with few-shot data, we can preserve previous scores but cannot provide enough samples to learn the corresponding weights for novel classes. For this reason, knowledge distillation~\cite{Tao_2020_CVPR} becomes a difficult problem in our case. Addressing this issue, we take advantage of a semantic word vector (word2vec \cite{Mikolov_NIPS_2013} or GloVe \cite{Jeffrey_Glove_2014}) which provides a semantic representation for each class as auxiliary knowledge. Being inspired by the literature on zero-shot learning~\cite{cheraghian2019zeroshot,Cheraghian_WACV_2020,cheraghian2019mitigating,tfvaegan-eccv20,latem-cvpr16,Rahman_2019_ICCV,ASD_ACCV}, given an image as input, we estimate the semantic word vectors for the input instead of directly predicting its class label. Then, we measure the similarity of the predicted word vectors with the word vectors from the set of possible class labels, followed by a softmax layer applied to the similarity values to get the final score of the classes. One key benefit of this approach is that adding new classes while training on novel tasks does not come with new weights to train because the model attempts to predict fixed-length word vectors as an intermediate representation. No matter how many classes are present during fine-tuning, the network continues with its previous task of estimating the word vectors. In this new set-up, the distillation loss can easily accommodate new classes. One challenge of this approach is to obtain a good alignment of visual and semantic word vectors of few-shot instances. To address this issue, we employ automatically assigned superclass information of classes to train multiple embedding modules in parallel after the backbone network. The set of superclasses is attained from the semantic word vector space representations of the base task, and are then held fixed for the novel classes that follow. We determine an embedding for each superclass during training such that each embedding sees only the superclass set of classes. Hence, given a novel class, there is a selection of embeddings that each may be more or less suited.
We employ an attention module~\cite{Fang_2019_ICCV} to merge multiple embedding outputs and a loss to train the alignment appropriately with few-shot instances. It helps the network not to overfit on the few-shot instances as well as not becoming biased to the base classes. Figure~\ref{fig:architecture_diff} describes the key differences between conventional works and our method. With our proposed approach, we successfully beat the current state-of-the-art~\cite{Tao_2020_CVPR} on MiniImageNet, CUB200, and the CIFAR100 datasets thanks to the combined effect of using the auxiliary semantic information from word vectors and knowledge distillation in concert.

In summary, the contributions of this paper are:
\textbf{(1)}~A semantically-guided knowledge distillation approach for few-shot class-incremental learning using semantic word vectors,
\textbf{(2)}~A new visual-semantic alignment strategy for few-shot class-incremental learning using automatically assigned superclass annotations,
\textbf{(3)}~Extensive experiments validating the approach on MiniImageNet, CUB200, and CIFAR100 while achieving new state-of-the-art results.

\section{Related work}

\noindent
\textbf{Incremental learning:} Incremental learning means learning from a sequence of data which appear over time. In the literature~\cite{DBLP:journals/corr/abs-1904-07734}, incremental learning techniques are categorized into three groups, task-incremental learning~\cite{Chaudhry_2018_ECCV,riemer2018learning,v.2018variational}, domain-incremental learning~\cite{pmlr-v70-zenke17a,NIPS2017_6892}, and class-incremental learning~\cite{8100070,castro:hal-01849366,Hou_2019_CVPR,Wu_2019_CVPR,simon2021geodesic}. In this paper, we are only concerned with the third category, class-incremental learning, as we consider a unified output where the task label is not available during test time. Rebuffi~\etal~\cite{8100070}  keeps an “episodic memory” of the samples and incrementally adapts the nearest-neighbor classifier for the novel tasks. Castro~\etal~\cite{castro:hal-01849366} proposed an end-to-end incremental learning method. In this method, a knowledge distillation loss is used to keep information about previously seen classes, and a classification loss is employed to learn the new classes. Hou \etal~\cite{Hou_2019_CVPR} introduced a novel approach for incrementally learning a unified classifier that reduces the imbalance between old and new classes by cosine similarity, which excludes the bias in the classifier. Wu \etal~\cite{Wu_2019_CVPR} proposed a method for large scale incremental learning, where they correct the bias in the output of the model with the help of a linear model. Simon \etal~\cite{simon2021geodesic} propose a novel approach to the arsenal of distillation techniques. They construct low-dimensional manifolds for previous and current responses and minimize the dissimilarity between the geodesic responses connecting the manifolds.

\noindent\textbf{Few-shot incremental learning:} There are not many works addressing the FSCIL setting. Tao \etal~\cite{Tao_2020_CVPR} proposed this setting for the first time. They utilize a neural gas (NG) network to learn and maintain the topology of the feature manifold produced by various classes. Specifically, they introduce a method that alleviates the forgetting of the old classes by stabilizing the topology of the NG and enhancing the representation learning for few-shot novel classes by expanding and modifying NG to novel training samples. There is another category, called dynamic few-shot learning (DFSL)~\cite{8578557,ren19incfewshot,xtranet,Jie_2021_CVPR}, which is similar to FSCIL. Some works call this setting incremental few-shot learning but for clarity, in this work, we call this setting DFSL. The only difference is that in DFSL, there are only two sequences of tasks, in comparison to FSCIL, which contains several tasks. The first task containing many training samples is called base task, and the second task containing only a few training samples is called a novel task. 
Gidaris \etal~\cite{8578557} proposed an attention-based method that generates a classifier for the novel task from the classifier of the base task.  Mengye \etal~\cite{ren19incfewshot} notes that the method of recurrent back-propagation can back-propagate within the optimization process and helps the learning of novel tasks. Yoon \etal~\cite{xtranet} proposed a method which obtains a task-adaptive representation for novel tasks based on the information provided from the base task by an attention module.

\noindent\textbf{Knowledge distillation:} Knowledge distillation is a well-known procedure that is employed in incremental learning to address catastrophic forgetting. Distillation loss was initially introduced to convey knowledge between separate neural networks \cite{44873}. Later, Li \etal \cite{8107520} used a distillation loss to preserve the knowledge of the old tasks while learning the new ones using a classification loss. Shmelkov~\etal~\cite{Shmelkov_2017_ICCV}~proposed a method where the embedding and the classifier are trained together without the need for keeping samples of the training data. Castro~\etal~\cite{Castro_2018_ECCV} introduced an end-to-end method which consists of a classification loss for learning novel tasks and a distillation loss to retain information of the old task. Zhang~\etal~\cite{9093365}~introduced an approach to train an individual network for the novel classes, and then merging this new network with the network based on previous classes using a double distillation objective. Zhao~\etal~\cite{9156766} employed knowledge distillation, at first, to keep the discrimination of old classes. Next, to further keep the balance between old and new classes, they introduced a method to refine the bias weights in the FC layer following the regular training. While it has been shown in~\cite{Tao_2020_CVPR} that regular knowledge distillation is not working well in the FSCIL setting, we offer, in this work, a method that enables the use of knowledge distillation for this purpose. 

\section{Method}

\subsection{Problem Formulation}


Suppose, there is a sequence of disjoint tasks $\mathcal{D} = \{ \mathcal{D}^{1}$, ...,  $\mathcal{D}^{T}$\}, where $\mathcal{C}^{t} = \left \{ c^{t}_{1},..., c^{t}_{m^{t}} \right \}$ is the set of classes in the task $\mathcal{D}^{t}$. Additionally, a set of $d$-dimensional semantic class embeddings for each class label in the task $\mathcal{D}^{t}$ defined as $\mathcal{S}^{t}$ is available during training. To be more specific, in the task $\mathcal{D}^{t}=\left \{ \left ( \textbf{x}^{t}_{i},l^{t}_{i},\textbf{s}^{t}_{i} \right ) \right \}_{i=1}^{|\mathcal{D}^{t}|}$, $\textbf{x}^{t}_{i}$ is the $i$-th sample, $l^{t}_{i}$ is its associated ground truth, and $\textbf{s}^{t}_{i}$ is its associated semantic representation. There are many training samples in the first task $\mathcal{D}^{1}$, termed the base task. However, in the following tasks $t>1$, termed the novel task, there are only a few training samples (5-shots per class) for each class.  It is essential to mention that the classes between all tasks are disjoint, \textit{i.e.}, $ \mathcal{C}^{i}\cap \mathcal{C}^{j}= \phi, \forall i,j \in \{1,\ldots , T \}$, where $i\neq j$.
The objective of our work is to incrementally train a model with a unified output, while only training samples of the $t$-th task is available at the $t$-th session. At test time, we expect the trained model on task $\mathcal{D}^{t}$ to predict the output for the current task and all the previous tasks \{$\mathcal{D}^{1}$, ... ,  $\mathcal{D}^{t-1}$\}.

\subsection{Knowledge Distillation}

\begin{figure}[!t]\centering
\includegraphics[width=1\linewidth,trim=0cm 0cm 0cm 0cm, clip]{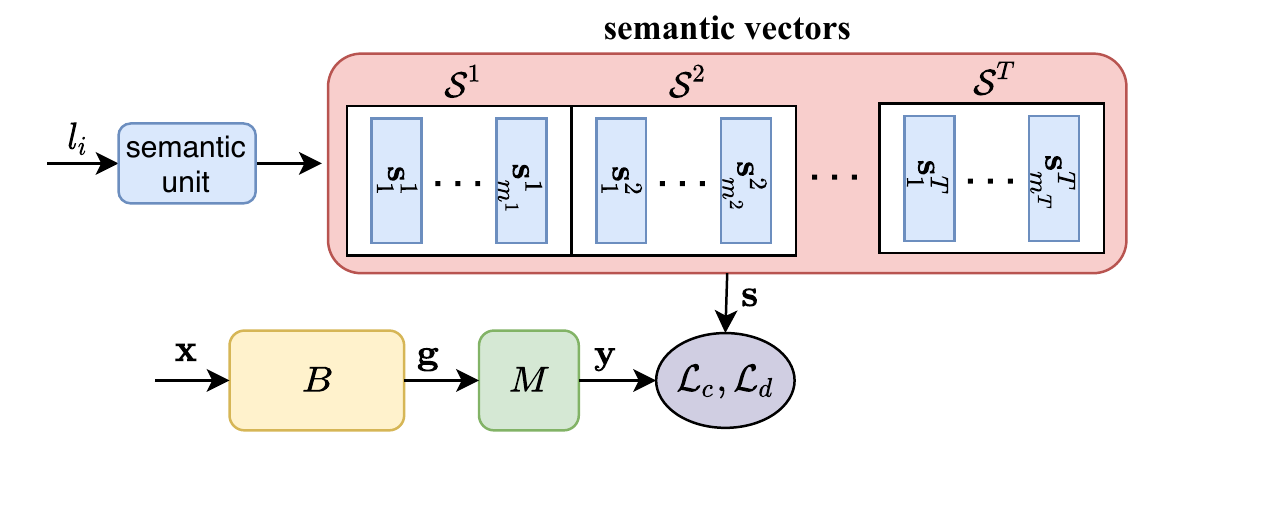}
\vspace{-9.5mm}
\caption{ Our simplified proposed architecture for knowledge distillation. In this design, the input image $\textbf{x}$ is forwarded into the backbone $B$ to extract a feature representation $\textbf{g} \in \mathbb{R}^{u}$. Then, the extracted feature ${\textbf{g}}$ is mapped into the semantic domain via a mapping module $M$ to form the estimation of the semantic vector $\textbf{y}\in \mathbb{R}^{d}$.}
\label{fig:architecture_base}
\vspace{-0.5cm}
\end{figure}
Knowledge distillation is a common approach \cite{8100070,castro:hal-01849366,Hou_2019_CVPR} for incremental learning to address the catastrophic forgetting. Even though it has shown promising results on incremental learning, it cannot be employed directly to few-shot class incremental learning (FSCIL) due to the imbalanced data and trade-off issues~\cite{Tao_2020_CVPR}. In this paper (see Figure \ref{fig:architecture_base}), we will show how to successfully take advantage of knowledge distillation in the FSCIL setting. To illustrate our proposed method for knowledge distillation, we use a simplified version of our proposed architecture. In this scheme, the input image $\textbf{x}$ goes into the backbone $B$, which is trained on only the first task $\mathcal{D}^{1}$, as we have many training samples in this task, and is kept frozen on other tasks. The output of the backbone $B$ is a feature representation $\textbf{g} \in \mathbb{R}^{u}$. Next, the mapping network $M$ is used to project the feature representation $\textbf{g}$ into the semantic domain, where the projected feature ${\textbf{y}}\in \mathbb{R}^{d}$ is aligned with its associated semantic representation $\textbf{s}\in \mathbb{R}^{d}$.  Lets assume we want to train the model with the task $\mathcal{D}^{t}$, which has $m$ or $|\mathcal{C}^t|$ classes, and the number of classes of the previous tasks are $n$. Also, we choose one representation for each class of the former tasks saved in a small memory $\mathcal{M}$, where each prototype  $\textbf{g}^{\mathcal{M}} \in \mathbb{R}^{u}$ is the average of all training samples from each class.
The aim is to map the input image $\textbf{x}_{i}$ into the semantic domain $\textbf{y}_{i}$ by the function $\textbf{y}_{i}=\kappa_{\theta}(\textbf{x}_{i})$, which consists all training parameters of the backbone $B$ and the mapping module $M$. We consider the cosine distance as the similarity between the projected feature $\textbf{y}_{i}$ and the semantic representation $\textbf{s}_{k}$, $d(\textbf{s}_{k},\textbf{y}_{i}) = cos(\textbf{s}_{k},\textbf{y}_{i})$. If the output of the classifier before adding the novel task $\mathcal{D}^{t}$ is $\textbf{d}{}' = \left [ d'(\textbf{s}_{1},\textbf{y}_{i}),...,d'(\textbf{s}_{n},\textbf{y}_{i})  \right ]$, and if the output of the classifier after adding the novel task is $\textbf{d} = \left [ d(\textbf{s}_{1},\textbf{y}_{i}),...,d(\textbf{s}_{n},\textbf{y}_{i}),d(\textbf{s}_{n+1},\textbf{y}_{i}), ..., d(\textbf{s}_{n+m},\textbf{y}_{i}) \right ]$, the distillation loss is defined as, 




\vspace{-3.5mm}
\begin{align}
\mathcal{L}_{d} =-\frac{1}{{N}_{c}} \sum_{i=1}^{{N}_{c}}\sum_{k=1}^{n}p_k log (q_k),
\label{eqn:distilation}
\end{align}
 \vspace{-1.5mm}
\begin{align}
p_k = \frac{e^{{-d{}'(\textbf{s}_{k},{\textbf{y}}_{i})/\tau }}}{\sum_{j=1}^{n}e^{-{d{}'(\textbf{s}_{j},{\textbf{y}}_{i})/\tau}}}~~,  & ~~~~~q_k = \frac{e^{{-d(\textbf{s}_{k},{\textbf{y}}_{i})/\tau}}}{\sum_{j=1}^{n}e^{-{d(\textbf{s}_{j},{\textbf{y}}_{i})/\tau}}}~~,\nonumber
\end{align}
where ${\textbf{y}}_{i}=M(\textbf{g}_{i})$, and {$\textbf{g}_{i} \in B(\textbf{x}^{t}_{i}) \cup \mathcal{M}$}.
Here, $\tau$ is the temperature scalar. The $\tau$ is set to 2 for all experiments. Also, $N_{c}$ is the number of samples in the task $\mathcal{D}^{t}$ and memory $\mathcal{M}$, i.e., $N_c = |\mathcal{D}^t| + |\mathcal{M}|$.

\begin{remark}
In this paper, we show that knowledge distillation can be used for few-shot class incremental learning. To this end, we use a semantic word vector in our pipeline as additional information. 
\end{remark}


Additionally, we employ a cross-entropy loss as a classification loss,


\vspace{-3.5mm}
\begin{align}
\mathcal{L}_{c} = -\frac{1}{N_{c}} \sum_{i=1}^{N_{c}}\sum_{k=1}^{n+m}\textbf{1}\left [ l_{i}==k \right ] log(\frac{e^{{-d(\textbf{s}_{k},{\textbf{y}}_{i})}}}{\sum_{j=1}^{n+m}e^{-{d(\textbf{s}_{j},{\textbf{y}}_{i})}}}),
\label{eqn:classification}
\end{align}
\noindent where $\textbf{1}\left [ \cdot \right ]$ is the indicator function.
Then, the total loss is defined as,
\vspace{-3.5mm}
\begin{align}
\mathcal{L} = \lambda_{1} \mathcal{L}_{d} + \lambda_{2}\mathcal{L}_{c},\nonumber
\label{eq:final_loss}
\end{align}
where $\lambda_{1}$ and $\lambda_{2}$ are used to control the effect of each loss in the final loss $\mathcal{L}$.

\begin{figure}[!t]
\includegraphics[width=1\linewidth,trim=0cm 0cm 0cm 0cm, clip]{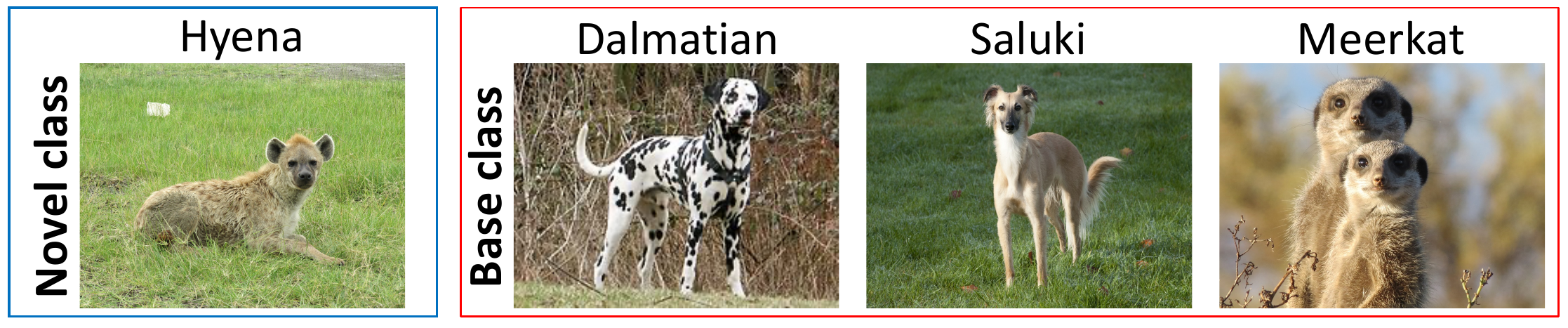}
\centering
\vspace{-1em}
\caption{\small For a novel class, Hyena, our model finds a set of base class (e.g., Dalmatian, Saluki, Meerkat) instances, which reside in close proximity in the semantic embedding space. The presence of shared semantics (e.g., face, body-shape, 4-feet, short-tail) between novel and base classes help to understand Hyena as a novel class and not to forget base classes.}
\label{fig:intuition}
\end{figure}

This approach does not add new parameters during the incremental learning stage. Instead, it only fine-tunes $M$ with newly available data incrementally. It helps the distillation process and learning without forgetting. Also, this approach reduces the data imbalance problem of few-shot learning. The network may not have had the opportunity to see the novel objects, however, newly discovered novel objects may very well share semantic properties with the objects it has already seen among the base classes.
For example, if hyena is a novel class, the network did not have any opportunity to see any hyena. However, many typical hyena attributes like `face', `body', etc., are seen by the network from other base class animals (e.g., Dalmatian, Saluki, Meerkat). It helps the network to understand hyena by reducing the data dependency. An intuitive illustration is shown in Figure \ref{fig:intuition}.

\subsection{Multiple embeddings for few shot tasks }

\begin{figure}[!t]\centering
\includegraphics[width=1\linewidth,trim=0cm 0cm 0cm 0cm, clip]{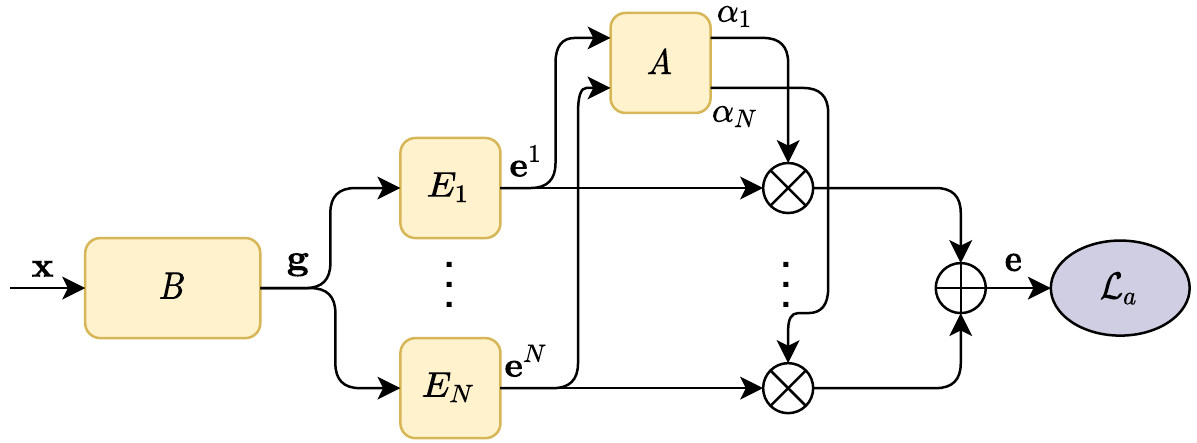}
\caption{Our proposed architecture is designed specifically for tasks with only a few training samples. The input image $\textbf{x}$ goes into the backbone $B$, which generates a global representation $\textbf{g} \in \mathbb{R}^{u}$. Then, the feature $\textbf{g}$ separates into several embedding modules $E$. The attention module $A$ is used to merge all of them to generate the final representation $\textbf{e} \in \mathbb{R}^{u}$. }
\label{fig:multiple_embedding}
\end{figure}

The main challenge with the tasks containing a few training samples is the overfitting issue, as it is difficult to learn the distribution of a task sufficiently with only a few training data. To this end, in our proposed method (see Figure~\ref{fig:multiple_embedding}), we generate multiple embeddings, where each is designed specifically for a group of classes. We use the word vector semantic to separate classes into several groups. We design these groups based on the classes that we have in the first task $\mathcal{D}^{1}$ because the first task contains more classes with many training samples. After that, we train each embedding $E_{i}$ using a cross-entropy loss based on the superclass labels obtained in the previous stage. 
Each embedding observes only part of the entire set of classes, which results in embeddings that are experts on a particular superclass.
For the novel tasks, we assign a superclass label to each class based on the cluster that we obtained in the previous stage.

\noindent
\textbf{Superclass:} The number of the embeddings is defined by the superclass knowledge obtained from the semantic word vector space. In the semantic space, there is a class embedding $\textbf{s}$ for each class. We apply $k$-means clustering of the semantic representations of the classes of the first task $\mathcal{D}^{1}$. In this way, similar classes fall into the same category. After applying $k$-means clustering, we assign a superclass label to each class $ \mathcal{R} = \left \{ {1},...,{N} \right \}$.
For the other tasks, which have only a few training samples, we use the cluster centers (obtained for the first task $\mathcal{D}^{1}$) to assign superclass labels to the classes in these tasks. To assign a superclass label to novel classes, we simply calculate the minimum Euclidean distance between the semantic vector of novel class and cluster centers.



\begin{figure*}[!t]\centering
\includegraphics[width=1\linewidth,trim=0cm 0cm 0cm 0cm, clip]{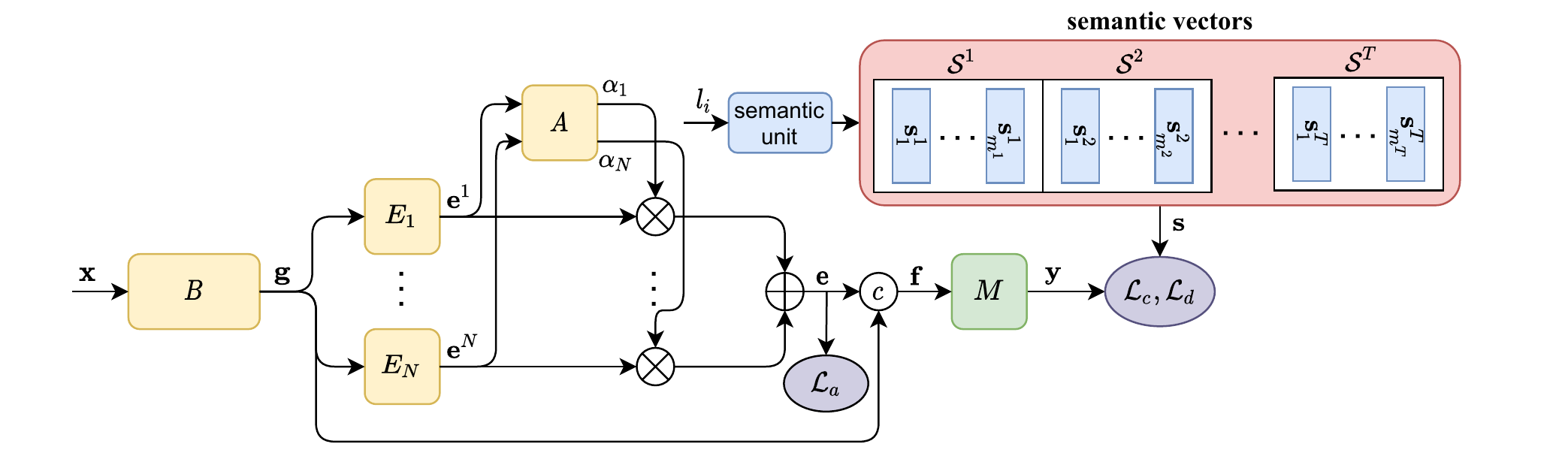}
\caption{ The proposed architecture. The image $\textbf{x}$ is forwarded to the Backbone $B$ to obtain the global feature representation $\textbf{g} \in \mathbb{R}^{u}$. The feature $\textbf{g}$ is fed into several embedding modules to get a representation $\textbf{e} \in \mathbb{R}^{u}$ based on the superclass information obtained from the word vector semantic space. After this stage, the feature vector $\textbf{f} \in \mathbb{R}^{2u}$ is obtained by concatenating the global and embedding representations. In the next stage, the $\textbf{f}$ is projected via a mapping module $M$ from the visual space into the semantic space ${\textbf{y}} \in \mathbb{R}^{d}$ to align the visual information with their associated semantic representation. }
\label{fig:architecture}
\vspace{-0.5cm}
\end{figure*}

\noindent
\textbf{Attention:} For the $i$-th sample in the task $\mathcal{D}^{t}$, the embedding representation $\textbf{e}_{i} \in \mathbb{R}^{u}$ is the weighted average of all private modules $\textbf{e}_{i}^{k} \in \mathbb{R}^{u}$ , $k \in \mathcal{R}$ , where a neural network determines weights, which is an attention $A$ module. Furthermore, the weights must sum to 1 to stay invariant to the number of private modules. Then, the final embedding representation is $\textbf{e}_{i} = \sum_{k=1}^{N}\alpha^{k} \textbf{e}_{i}^{k}$, where $\alpha^{k}$ is, 



\noindent
where

\vspace{-3.5mm}
\begin{align}
\alpha^{k}=\frac{e^{{\textbf{w}^{T}tanh(\textbf{V}(\textbf{e}_{i}^{k})^{T})}}}{\sum_{j=1}^{N}e^{{\textbf{w}^{T}tanh(\textbf{V}(\textbf{e}_{i}^{j})^{T})}}}.
\end{align}

\vspace{-1.5mm}

\noindent
Here, $\textbf{w}\in \mathbb{R}^{L \times 1}$ and $\textbf{V}\in \mathbb{R}^{L \times d}$ are trainable parameters.


\begin{algorithm}[!t]
\caption{Training procedure of the proposed method}\label{euclid}
\begin{algorithmic}[1]

\Function{train}{$\mathcal{D}$} 

\State Hyperparameters: $\lambda_{1}$, $\lambda_{2}$, $\lambda_{3}$, $N$ 
\State $\mathcal{M} \leftarrow \left \{  \right \}$
\State Train the backbone $B$ on the first task $\mathcal{D}^{1}$ 



\State Apply $k$-means clustering, where $k=N$, on base \hspace*{5mm} semantic vectors and assign a superclass label

 
\State Train $N$ embedding $E_{i}$ (Fig.~\ref{fig:multiple_embedding}) on the base task \hspace*{5mm} $\mathcal{D}^{1}$ using superclass labels as cluster identity 

\algorithmiccomment{Train the final architecture (Fig.~\ref{fig:architecture})}
\For{$h=1$ to $epochs$}
     
\State Calculate the classification loss using Eq~\ref{eqn:classification} 
\State Calculate the attention loss using Eq~\ref{eqn:attention} 
\State $\mathcal{L}=\lambda_{1}\mathcal{L}_{c} + \lambda_{3}\mathcal{L}_{a}$

\State Backpropagate and update $A$, and $M$

\EndFor

\State $\mathcal{M}\leftarrow \text{UPDATEMEMORY}(\mathcal{D}^{1}$, $\mathcal{M}$, $\mathcal{C}^{1})$

\For{$t=2$ to $T$}
\State Assign superclass labels to task $\mathcal{D}^{t}$ using \hspace*{10mm}  cluster centers obtained for the base task
\For{$h=1$ to $epochs$}
     
\State Calculate the classification loss using Eq~\ref{eqn:classification} 
\State Calculate the distillation loss using Eq~\ref{eqn:distilation} 
\State Calculate the attention loss using Eq~\ref{eqn:attention} 
\State $\mathcal{L}=\lambda_{1}\mathcal{L}_{c} +\lambda_{2}\mathcal{L}_{d}+ \lambda_{3}\mathcal{L}_{a}$

\State Backpropagate and update $A$, $M$, and $E$

\EndFor

\State $\mathcal{M}\leftarrow \text{UPDATEMEMORY}(\mathcal{D}^{t}$, $\mathcal{M}$, $\mathcal{C}^{t}_{m^{t}})$

\EndFor

\EndFunction

\Function{updatememory}{$\mathcal{D}^{t}$, $\mathcal{M}$, $\mathcal{C}^{t}_{m^{t}}$}

\For{$c=1$ to $ | \mathcal{C}^{t}_{m^{t}} |$}
 \State \hspace*{-1mm}Calculate a prototype $\textbf{g}_{c}^{\mathcal{M}}$ for each class by \hspace*{9mm} averaging of all training samples from each class    
\State $\mathcal{M}\leftarrow \mathcal{M}\cup (\textbf{g}_{c}^{\mathcal{M}},\textbf{l}^{t}_{c})$
\EndFor
\State \textbf{return} $\mathcal{M}$
\EndFunction

\end{algorithmic}

\label{alg}
\end{algorithm}

\noindent
\textbf{Training:} We train the attention modules $ A $ so that the final representation $\textbf{e}$ becomes similar to the corresponding embedding module given the associated superclass label. To this end, we use the following loss function,


\vspace{-3.5mm}
\begin{align}
\mathcal{L}_{a} = \frac{1}{N_{t}}\sum_{i=1 }^{N_{t}}\frac{e^{{-d(\textbf{e}_{i},\textbf{e}_{i}^{k})}}}{ \sum_{j=1}^{N}e^{{-d(\textbf{e}_{i},\textbf{e}_{i}^{j})}}  } \label{eqn:attention}
\end{align}
\noindent where the superclass label for $i$-th sample is $k \in \mathcal{R}$, and  $N_{t}$ is the number of samples in the task $\mathcal{D}^{t}$. 
This approach helps to not overfit the network on only a few novel class data. Multiple embeddings specialized on related classes, belonging to the same superclass, describe the novel instances. 
Combining multiple embedding features and the global feature enables a strong generalization when classifying both base and novel classes.

\subsection{Model Overview}

The final proposed architecture is shown in Figure~\ref{fig:architecture}. The input image $\textbf{x}$ goes into the backbone $B$, which is pretrained on the first task $D^{1}$ with a cross-entropy loss, to generate the global feature representation $\textbf{g} \in \mathbb{R}^{u}$. 
When training the other tasks $\mathcal{D}^{t}, t>1$, the backbone network $B$ is kept frozen to prevent overfitting to classes with only a few training samples. Then, the extracted feature $\textbf{g}$ is fed into $N$ embedding modules $E_i,~i = 1, \ldots, N$, where they are trained on the first task based on the superclass information and are updated for the novel tasks.
To this end, the output of all embedding $\textbf{e}^{k} \in \mathbb{R}^{u}$ are fused based on the weights generated by an attention module $A$. Then, the global feature $\textbf{g}$ is concatenated with the feature generated from the embedding modules $\textbf{e}$ to form the feature ${\textbf{f}} \in \mathbb{R}^{2u}$. In the next stage, the feature ${\textbf{f}}$ is projected from the visual domain to the semantic domain by the mapping module $M$ to form feature ${\textbf{y}}\in \mathbb{R}^{d}$, where it is aligned with its associated semantic representation $\textbf{s} \in \mathbb{R}^{d}$. In order to train our proposed architecture, we use the following loss function,

\vspace{-3.5mm}
\begin{align}
\mathcal{L} = \lambda_{1} \mathcal{L}_{c} + \lambda_{2}\mathcal{L}_{d}+ \lambda_{3}\mathcal{L}_{a},
\end{align}
where $\lambda_{1}$, $\lambda_{2}$, and $\lambda_{3}$ are used to control the effect of each term in the final loss function. The pseudo code of the training procedure is shown in Algorithm~\ref{alg}.

\section{Experiments}

This section contains two parts. In the first part, we evaluate our method on FSCIL~\cite{Tao_2020_CVPR}, and we conduct a set of ablation studies to investigate the recommended approach. Next, we investigate the dynamic few-shot learning~\cite{8578557} (DFSL) setting to demonstrate the capability of our proposed method in a different setting. 

\subsection{Experiments on FSCIL}

\begin{table*}[]
\centering
\caption{CUB200 results with ResNet18 based on the 10-{way} 5-{shot} setting.}
\scalebox{.9}{
\begin{tabular}{lcccccccccccc}
\hline
\multirow{2}{*}{Method} & \multicolumn{11}{c}{Sessions}  \\ \cline{2-12}
 & 1 & 2 & 3 & 4 & 5 & 6 & 7 & 8 & 9 & 10 & 11   \\ \hline
iCaRL~\cite{8100070} & 68.68 & 52.65 & 48.61 & 44.16 & 36.62 & 29.52 & 27.83 & 26.26 & 24.01 & 23.89 & 21.16  \\
EEIL~\cite{castro:hal-01849366} & 68.68 &53.63  & 47.91 & 44.20 & 36.30 & 27.46 & 25.93 & 24.70 & 23.95 & 24.13 & 22.11   \\
NCM~\cite{Hou_2019_CVPR} & 68.68 & 57.12 & 44.21 & 28.78 & 26.71  &25.66  & 24.62 & 21.52 & 20.12 &20.06  & 19.87   \\
AL-MML~\cite{Tao_2020_CVPR} & 68.68 & \textbf{62.49} & {54.81} & {49.99} & {45.25} & 41.40 & 38.35 & 35.36 & 32.22 & 28.31  & 26.28   \\
Ours & 68.23 & 60.45 & \textbf{55.70} & \textbf{50.45} & \textbf{45.72} & \textbf{42.90} & \textbf{40.89} & \textbf{38.77} & \textbf{36.51} & \textbf{34.87} & \textbf{32.96}   \\ \hline

\end{tabular}
\label{table:CUB}
}
\end{table*}


\noindent\textbf{Datasets:} We evaluate our proposed method on three well-known datasets, MiniImageNet~\cite{NIPS2016_6385}, CUB200~\cite{WahCUB_200_2011}, and CIFAR100~\cite{CIFAR-100}. MiniImageNet contains 100 classes, where each class include 500 training samples and 100 testing samples. The size of each image is $84\times84$. CUB200 contains 200 fine-grained classes, separated into 6000 training images, and 6000 testing images. The image size in this dataset is $224\times224$. CIFAR100 includes 100 classes, where each class has 600 images, separated into 500 training images and 100 test images. Each image has a size of $32\times32$. In this paper, we follow the setting proposed by~\cite{Tao_2020_CVPR}. In this setting, for MiniImageNet and CIFAR100, $60$ and $40$ are selected as the number of base and novel classes. For novel classes, a 5-\textit{way} 5-\textit{shot} setting is considered. There are nine sessions for MiniImageNet and CIFAR100 datasets (1 base session + 8 novel sessions). For the CUB200 dataset, a 10-\textit{way} 5-\textit{shot} setting is considered, where 100 classes are selected as base classes, and the remaining 100 classes are split into 10 sessions.

\noindent\textbf{Semantic Features:} We use unsupervised word vectors trained on an unannotated text corpus as a class semantic embedding.  For MiniImageNet, CUB200, and CIFAR100, we employ 1000, 400, and 300  dimensional word2vec~\cite{Mikolov_NIPS_2013}, respectively. For the ablation study, we also use the 300 dimensional GloVe~\cite{Jeffrey_Glove_2014} for the CUB200 dataset. 

\noindent\textbf{Validation:} To find hyperparameters, we conducted a grid search. The ranges we consider, $\lambda_{1}$,$\lambda_{2}$, $\lambda_{3} \in [0,2]$, \# of embedding modules $E \in [1,10]$, and the temperature $\tau \in [0,5]$ value. We split the training set into two sets: base set, which consists 60\% of the training classes, and validation set which consists rest of the classes added incrementally. 

\noindent\textbf{Implementation details\footnote{Code is available at: \url{https://github.com/ali-chr/Semantic-aware-Knowledge-Distillation-for-Few-ShotClass-Incremental-Learning}}:} We employ ResNet18~\cite{He_2016_CVPR} for the backbone $B$, where features of the input image are derived from the final pooling layer with 512 dimensions. The backbone $B$ is trained on the first/base task, and kept fixed for the following tasks. For the embedding modules $E_{i}$, we use a fully connected layer with 512 dimensions. Moreover, we utilize a few fully connected layers for the attention module $A$. Ultimately, for the mapping module $M$, we used three fully-connected layers with 512, 728, and $d$, which is the dimension of the semantic word vector, hidden units, where all layers have a ReLU function. In all experiments, we utilize the Adam optimizer~\cite{Article40}, where the learning rate and batch size were set to 0.001 and 128, respectively. The number embedding modules for MiniImageNet, CUB200, and CIFAR100 are selected as 3,5, and 3 respectively. Also, the value of $\lambda_{1}$, $\lambda_{2}$, and $\lambda_{3}$ are 0.7 , 1.1 , and 0.6 respectively for all datasets. Since we use semantic word vectors in our pipeline, we get slightly different result on the first task in comparison to the setting proposed by~\cite{Tao_2020_CVPR}.

\subsubsection{Results}
\begin{figure}[!t]
\includegraphics[width=1\linewidth,trim=0cm 0cm 0cm 0cm, clip]{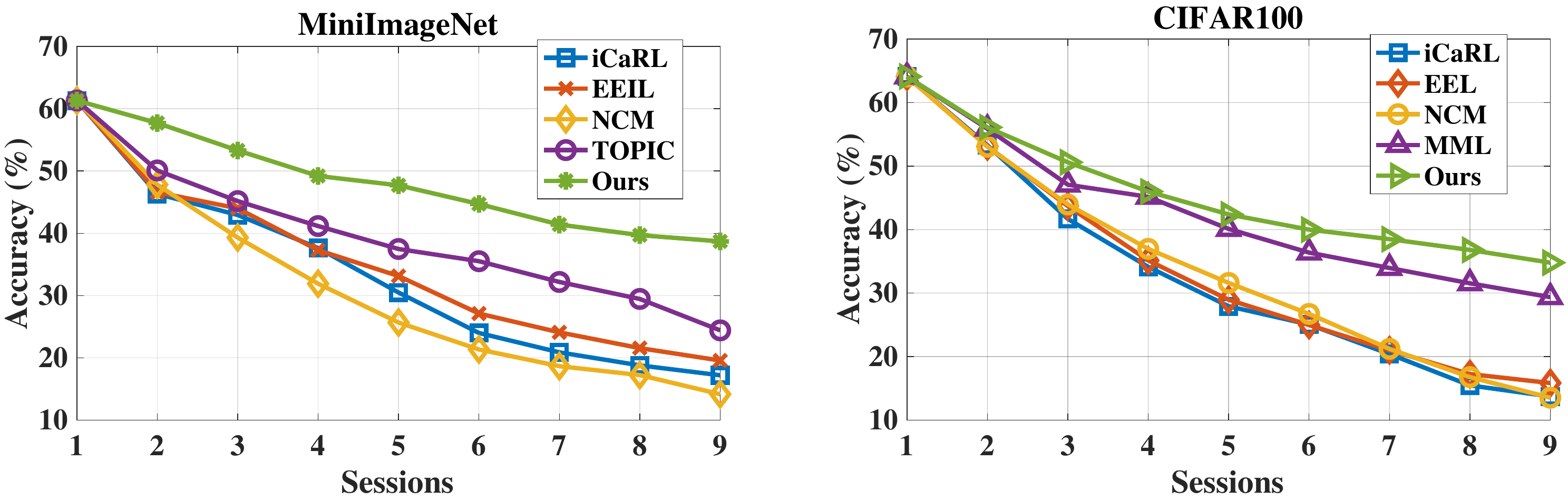}
\centering
\caption{
Results on MiniImageNet and CIFAR100 based on the 5-way 5-shot FSCIL setting.
}
\label{fig:MiniImageNet_CIFAR}
\vspace{-0.5cm}
\end{figure}

In this part, we compare our proposed method with state-of-the-art~\cite{8100070,castro:hal-01849366,Hou_2019_CVPR,Tao_2020_CVPR} on the 5/10-way and 5-shot setting. Figure~\ref{fig:MiniImageNet_CIFAR} presents our results on MiniImageNet and CIFAR100 using the 5-{way} 5-{shot} setting. Moreover, we show the result on the CUB200 dataset in Table~\ref{table:CUB}. Overall, on these three datasets, our method beats all state-of-the-art methods. As additional sequences of new tasks arrive, our approach shows its advantages to the other methods. To be more specific, in MiniImageNet, in the last session, we get 39.04\% accuracy, while the second-best one (TOPIC) achieves 24.42\% accuracy which demonstrates that our method surpasses the state-of-the-art by a large margin (more than 14\%). On CIFAR100, our method reaches the absolute accuracy of 34.80\%, while the second-best (TOPIC) one accuracy is 29.37\%. Also, on CUB200, our approach achieves 32.96\% in the last session, which is superior to the other approaches.

\subsubsection{Ablation study}

\begin{figure}[!t]
\includegraphics[width=1\linewidth,trim=0cm 0cm 0cm 0cm, clip]{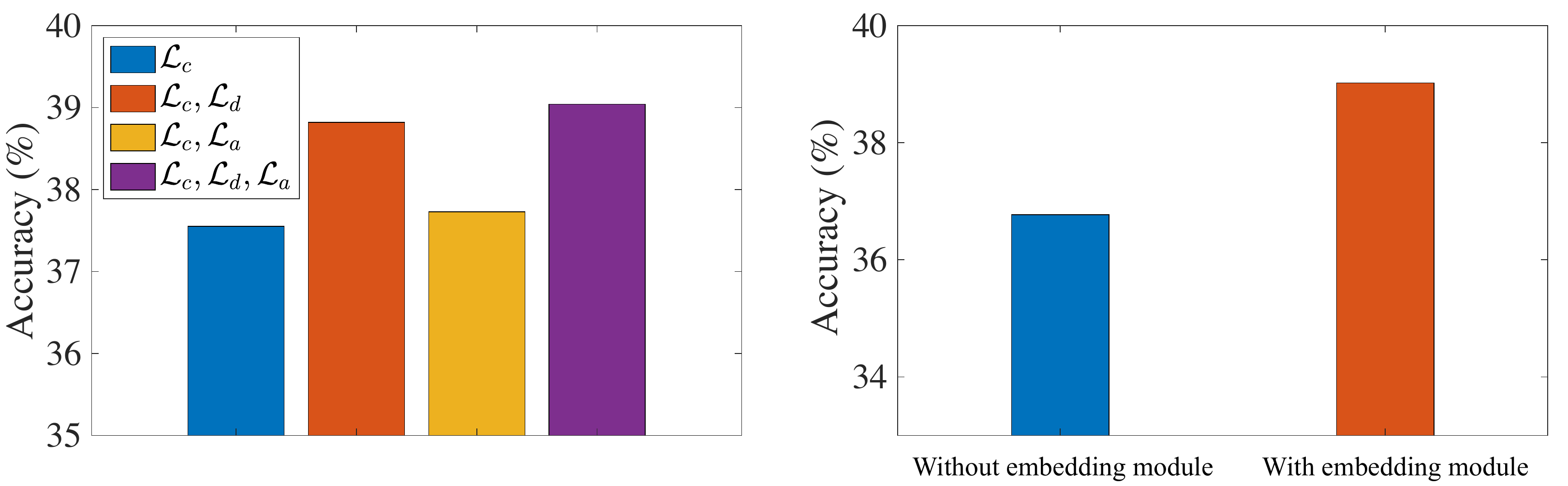}
\centering
\caption{
\small (Left) The influence of $\mathcal{L}_{d}$ and $\mathcal{L}_{a}$ losses, and (Right) the impact of using multiple embedding 
}
\label{fig:embedding_impact}
\end{figure}

\noindent\textbf{Impact of loss function and embedding module:} In this part, we report the individual effect of the distillation loss $\mathcal{L}_{d}$ and attention loss $\mathcal{L}_{a}$  functions in the total loss. As can be seen in Figure~\ref{fig:embedding_impact}, where only the accuracy of the last session is shown, $\mathcal{L}_{d}$ is more effective than $\mathcal{L}_{a}$, as $\mathcal{L}_{d}$ helps the model to remember the previously seen tasks, while $\mathcal{L}_{a}$ helps the model to generate a richer feature representation for the novel tasks, which have only a few training samples.

We also evaluate the effect of multiple embedding modules in our proposed algorithm. To understand the influence of these in our design, we utilise a baseline architecture which does not include the multiple embedding module and has only the backbone $B$. As presented in Figure~\ref{fig:embedding_impact}, our proposed method exceeds the baseline, which indicates the value of the embedding module inside our architecture.

\begin{figure*}[!t]
\includegraphics[width=.9\linewidth,trim=0cm 0cm 0cm 0cm, clip]{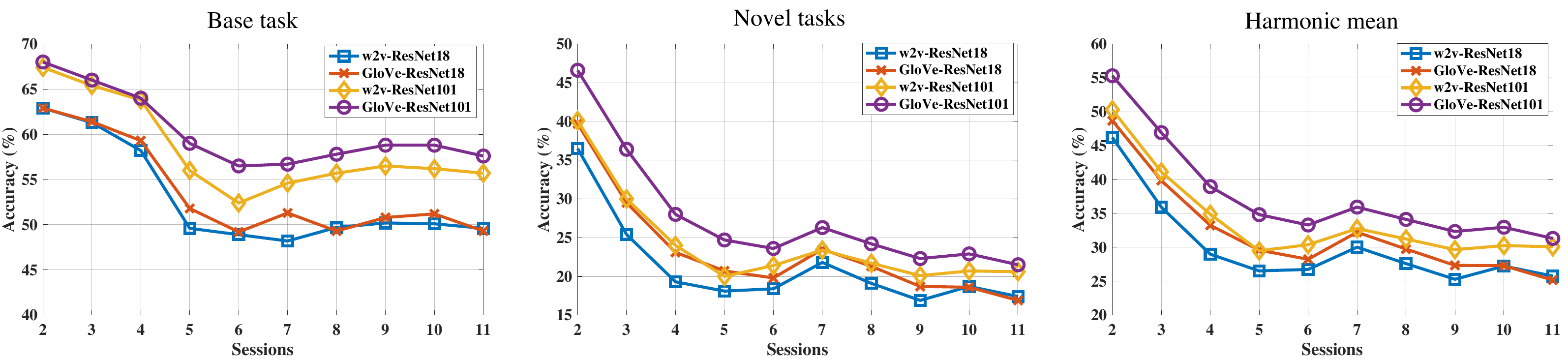}
\centering
\caption{
\small The results of our method on base (left), novel tasks (middle), and their harmonic mean (right) in different incremental sessions.}
\label{fig:old_novel_HM}
\end{figure*}


\noindent\textbf{{Impact of different backbones and semantic information:}} In this ablation study, we evaluate the influence of the alternative semantic word vector, GloVe. We also conduct this analysis using two backbones, ResNet18 and ResNet101, to understand the effect of different backbones. The accuracy of our approach is evaluated on test samples of the base task and the novel tasks.
The accuracy of the base task $Acc_{b}$ is the performance of our model on the first/base task $\mathcal{D}^{1}$, and the accuracy of the novel task $Acc_{n}$ is considered the performance on the test samples of the current task and all previous novel tasks $\left \{ \mathcal{D}^{2},...,\mathcal{D}^{t} \right \}$. To evaluate the contribution of the base and novel instances in the final accuracy, we also report the Harmonic Mean (HM)~\cite{10.1007/978-3-319-46475-6_4} of the accuracy of the base and novel classes.


The values for $Acc_{b}$, $Acc_{n}$, and HM are shown in Figure~\ref{fig:old_novel_HM}. As can be seen, the combination of GloVe and ResNet101 forget less of the base classes, which is followed by better learning on novel classes. As a result, it provides a greater HM value. It reveals that having a different type of semantic vector is valuable in our suggested pipeline. Additionally, a deeper backbone (ResNet101) is useful in FSCIL as it generates more valuable feature representations. We present our results starting from session 2 to 11 because there is no novel task in session 1.  

\begin{table}[]
\centering
\caption{miniImageNet 64+5-way results}
\scalebox{0.75}{
\begin{tabular}{lcccc}
\hline
\multirow{2}{*}{Method} & \multicolumn{2}{c}{1-shot} & \multicolumn{2}{c}{5-shot} \\
 & accuracy & $\Delta$ & accuracy & $\Delta$ \\ \hline
 Imprint~\cite{8578708} & 41.34 $\pm$ 0.54\% & -23.79\% & 46.34 $\pm$ 0.54\% &  -25.25\%  \\
LwoF~\cite{8578557} & 49.65 $\pm$ 0.64\% & -14.47\% & 59.66 $\pm$ 0.55\% & -12.35\%   \\
AA~\cite{ren19incfewshot} &  54.95 $\pm$ 0.30\% & {-11.84}\% & 63.04 $\pm$ 0.30\% & -10.66\% \\
XtarNet~\cite{xtranet} & {55.28 $\pm$ 0.33\%} & -13.13\% & {66.86 $\pm$ 0.31\% }&  {-10.34\%} \\
Ours & \textbf{58.07 $\pm$ 0.27\%} & \textbf{-10.83\%} &\textbf{68.03 $\pm$ 0.44\%} & \textbf{-8.25\%}  \\ \hline
\end{tabular}}
\label{table:DFSL}
\end{table}

\begin{figure}[!t]
\centering
\includegraphics[width=1\linewidth,trim=0cm 0cm 0cm 0cm, clip]{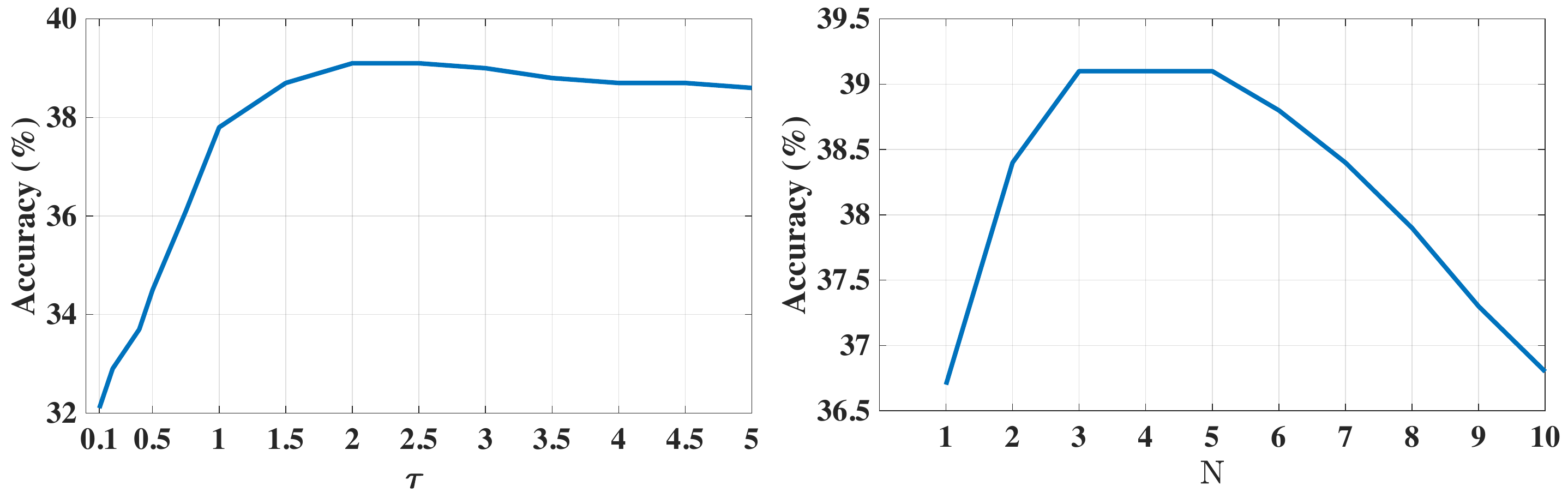}
\vspace{-2em}
\caption{
Impact of \textbf{(left)} Temperature, $\tau$ and \textbf{(right)} number of Supercluster ($N$) on the last task of MiniImageNet.
}
\label{fig:tau_N}
\vspace{-0.6cm}
\end{figure}

\noindent\textbf{Impact of Temperature, $\tau$:} We notice that the method is robust to $\tau$ in a wide range. Figure~\ref{fig:tau_N} \textbf{(left)} shows the effect of $\tau$ on MiniImageNet on the last task. Increasing $\tau$ is helpful for the forgetting issue until $\tau=2$ by balancing the contribution of $p_k$ and $q_k$, after that the accuracy decreases.

\noindent \textbf{Impact of number of Supercluster ($N$):} In the Figure~\ref{fig:tau_N} \textbf{(right)}, we present the performances on MiniImageNet on the last task when $N$ varies. When $N$ is very small (\eg, 1 or 2), it seems that our method only extracts global information, merging semantics of different super-categories, thus failing to encode local information and subtle differences within each super-category. Conversely, when $N$ is very high (\eg, $\geq 7$), fewer classes fall into the same super-category, which leads to indistinguishable information. As we concatenate both global and local information, with high $N$, local information does not help. Thus, using $N=10$, we got a similar performance to only global, $N=1$. Empirically, $3 \leq N \leq 5$ both global and local information work together for MiniImageNet.

\subsection{Experiments on DFSL}

To further evaluate our proposed method, we apply our design to the DFSL setting. As discussed in the related work section, DFSL is similar to FSCIL, though the only difference is that we have only two sequences of tasks in DFSL whereas there are various sequences of tasks in FSCIL. In this section, we use the setting introduced by~\cite{8578557}. In this experiment, we use miniImageNet, which is split into two tasks, a base task and a novel task. The base task consists of 64 classes, while the novel task has five classes which are randomly selected in an episode way from 20 classes. It is necessary to mention that the classes in the base and novel tasks are disjoint. In each episode, the training (support) set is built by selecting 1 or 5 examples of the novel classes, representing a 1-shot or 5-shot scenario. The test (query) set consists of examples of the base and novel tasks. In this experiment, we use ResNet12 as adopted in~\cite{mishra2018a} as the backbone. The forgetting in the DFSL setting is calculated as the gap between joint and individual performances for base and novel classes. Individual performances for the base/novel task is calculated when only the base/novel classifier is used. $\Delta_{b}$ and $\Delta_{n}$ are used to indicate the gap between the base and novel tasks, respectively. The average of these gaps is represented as $\Delta = (\Delta_{b}+\Delta_{n})/2$, which show the amount of forgetting, 

We compare our method with state-of-the-art methods in Table~\ref{table:DFSL}. As can be seen, our method outperforms all other methods on joint accuracy and forgetting $\Delta$ on both the 1-shot and 5-shot settings.




\section{Conclusion}


We proposed a semantic-aware knowledge distillation method for few-shot class incremental learning (FSCIL). Due to a limited amount of training data for the novel classes, the knowledge distillation technique as previously used did not work well in this problem domain. In this paper, using auxiliary information from class semantics (word vectors), we propose a new FSCIL method where knowledge distillation can indeed perform learning without forgetting. Moreover, we offer an attention mechanism based on multiple embedding representations of visual data to describe the novel classes that also demonstrates better generalization. Three well-known datasets, MiniImageNet, CUB200, and CIFAR100, are used to show that class semantics can be a useful source of information for knowledge distillation. We outperform state-of-the-art methods of FSCIL by a large margin.

{\small
\bibliographystyle{ieee_fullname}
\bibliography{egbib}
}

\end{document}